\documentclass[letterpaper]{article} 
\usepackage{aaai2026}  
\usepackage{times}  
\usepackage{helvet}  
\usepackage{courier}  
\usepackage[hyphens]{url}  
\usepackage{graphicx} 
\usepackage{natbib}  
\usepackage{caption} 
\usepackage{algorithm}
\usepackage{algorithmic}
\usepackage{booktabs}
\usepackage{multirow}
\usepackage{pifont}
\usepackage{amsmath}   
\usepackage{amssymb} 

\usepackage{newfloat}
\usepackage{listings}
\DeclareCaptionStyle{ruled}{labelfont=normalfont,labelsep=colon,strut=off} 
\floatstyle{ruled}
\newfloat{listing}{tb}{lst}{}
\floatname{listing}{Listing}
\title{Exploring Surround-View Fisheye Camera 3D Object Detection}
\author{
    Changcai Li\textsuperscript{\rm 1,2},
    Wenwei Lin\textsuperscript{\rm 1},
    Zuoxun Hou\textsuperscript{\rm 3}, 
    Gang Chen\textsuperscript{\rm 1}\thanks{Corresponding authors.}, \\
    Wei Zhang\textsuperscript{\rm 2}\footnotemark[1],
    Huihui Zhou\textsuperscript{\rm 2},
    Weishi Zheng\textsuperscript{\rm 1}
}
\affiliations{
    \textsuperscript{\rm 1}Sun Yat-sen University\\
    \textsuperscript{\rm 2}Pengcheng Laboratory\\
    \textsuperscript{\rm 3}Beijing Institute of Space Mechanics and Electricity \\

    cheng83@mail.sysu.edu.cn, zhangwei1213052@126.com
}

\usepackage{bibentry}

\begin{document}

\maketitle

\begin{abstract}
In this work, we explore the technical feasibility of implementing end-to-end 3D object detection (3DOD) with surround-view fisheye camera system.
Specifically, we first investigate the performance drop incurred when transferring classic pinhole-based 3D object detectors to fisheye imagery. To mitigate this, we then develop two methods that incorporate the unique geometry of fisheye images into mainstream detection frameworks: one based on the bird's-eye-view (BEV) paradigm, named FisheyeBEVDet, and the other on the query-based paradigm, named FisheyePETR.
Both methods adopt spherical spatial representations to effectively capture fisheye geometry.
In light of the lack of dedicated evaluation benchmarks, we release Fisheye3DOD, a new open dataset synthesized using CARLA and featuring both standard pinhole and fisheye camera arrays.
Experiments on Fisheye3DOD show that our fisheye-compatible modeling improves accuracy by up to 6.2\% over baseline methods.
\end{abstract}

\begin{links}
    \link{Code}{https://github.com/weiyangdaren/Fisheye3DOD}
\end{links}

\section{Introduction}

\begin{figure}[tbp]
	\centering
	\includegraphics[width=1\linewidth]{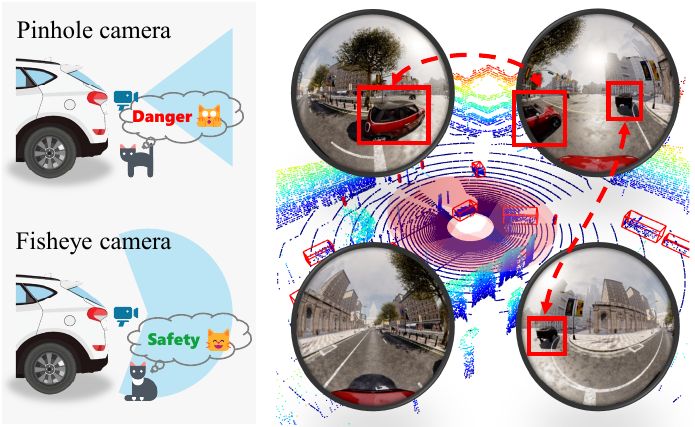}
	\caption{\textbf{Left:} The pinhole camera setup has blind spots in the near field, whereas the fisheye camera provides enhanced coverage for improved safety. \textbf{Right:} The same object is captured from multiple fisheye viewpoints.
	}
	\label{fig:image1}
\end{figure}

Reliable \( 360^\circ \) perception is vital for autonomous systems like self-driving vehicles and various robots.
Surround-view fisheye cameras enable this via a compact multi-camera setup, as shown in Figure~\ref{fig:image1} (\textbf{Right}), where four ultra-wide lenses (each exceeding \( 180^\circ \) field of view (FoV)) capture the full surroundings seamlessly.
Existing fisheye-based works focus on depth estimation~\cite{ won2020end, xie2023omnividar} and segmentation~\cite{deng2019restricted, playout2021adaptable}, but 3DOD—a crucial task for dynamic obstacle avoidance—remains unexplored.

Compared to current vision-centric 3D perception systems that mainly use multi-pinhole-camera setups~\cite{caesar2020nuscenes, sun2020scalability}, fisheye camera-based solutions offer several distinctive advantages.
\textbf{First}, because of regulatory requirements, such as the 2018 U.S. mandate to prevent reversing accidents through the use of rear-view fisheye cameras~\cite{sunstein2019rear}, modern mass-produced vehicles are widely equipped with such cameras (e.g., BMW~\cite{hughes2009wide}), as illustrated in Figure~\ref{fig:image1} (\textbf{Left}).
This allows direct use of the pre-installed sensors for perception, avoiding costly retrofits needed by pinhole systems.
\textbf{Second}, compared to methods~\cite{ge2023metabev, yan2023cross, xie2025benchmarking} that rely on algorithms to improve robustness against sensor failures, a surround-view fisheye system inherently provides physical redundancy via overlapping FoV. As shown in Figure~\ref{fig:image1} (\textbf{Right}), this overlap captures the same object from multiple viewpoints to enhance reliability.
\textbf{Finally}, the ultra-wide FoV suits scenarios with constrained space or cost-sensitive deployment, such as indoor robotics or surveillance. Standard pinhole setups usually require more cameras for similar coverage (e.g., nuScenes~\cite{caesar2020nuscenes} uses six, Tesla autopilot~\cite{tatarek2017functionality} uses eight).

Despite their advantages, fisheye cameras pose inherent challenges that can degrade detection performance. In particular, the non-linear projection compresses objects into very few pixels, making them harder to detect reliably. This issue is fundamental to fisheye imagery and motivates the following Research Questions (\textbf{RQs}): \textit{1) how much accuracy is lost when transferring pinhole-based detectors to fisheye images}, and \textit{2) how can the transfer be made more effective}?
Due to the absence of a unified benchmark, prior studies~\cite{plaut20213d, yogamani2024daf, yogamani2024fisheyebevseg} mostly focus on evaluating methods within their respective imaging models—either on pinhole only or on fisheye only—without directly comparing the two. As a result, these \textbf{RQs} remain open.
Answering these questions is vital both for advancing fisheye 3D perception and for guiding industrial design and sensor choices, where camera trade-offs affect cost, coverage, and reliability.

To address the \textbf{RQ1}, we create \textbf{Fisheye3DOD}, a synthetic dataset based on the CARLA simulator~\cite{dosovitskiy2017carla}. Fisheye3DOD provides synchronized multi-view data from both six surround pinhole cameras and four fisheye cameras under identical scenarios, enabling a direct and fair comparison between pinhole and fisheye imaging models for 3DOD task.
Then, we leverage the Fisheye3DOD dataset to systematically evaluate representative pinhole-based 3D detectors on fisheye data after rectification, including both perspective and cylindrical projections.

To address the \textbf{RQ2}, we introduce spherical back-projection at the feature level to resolve the geometric incompatibilities between fisheye and pinhole models. 
Current pinhole-based 3D detectors, such as BEVDet~\cite{huang2021bevdet} with explicit BEV construction or PETR~\cite{liu2022petr} with implicit 3D query encoding, rely on perspective projection assumptions that are incompatible with fisheye optics' nonlinear distortions.
To overcome this limitation, we propose two geometry-aligned frameworks that integrate spherical back-projection into mainstream detection pipelines.
Specifically, we introduce \textbf{FisheyeBEVDet} and \textbf{FisheyePETR}, both of which project image features onto a spherical equirectangular representation.
Then, FisheyeBEVDet adapts the BEV-based architecture by performing depth reasoning in spherical coordinates.
FisheyePETR, on the other hand, employs spherical ray-based positional encodings to enhance projected features, which then interact with object queries.
These two designs address the unique geometric distortions of fisheye cameras while maintaining compatibility with established 3D detection paradigms.

Experiments on Fisheye3DOD demonstrate that directly transferring pinhole-based detectors to rectified fisheye images leads to a significant accuracy drop. 
Through end-to-end optimization with spherical modeling at the feature level, our proposed FisheyeBEVDet and FisheyePETR effectively mitigate the performance gap, improving detection accuracy by 4.5 and 6.2 points, respectively.

In summary, our key contributions are as follows:
\begin{itemize}
	\item To our knowledge, we present the first systematic and quantitative study comparing 3D perception performance between pinhole and fisheye imaging.
	\item We introduce Fisheye3DOD, a benchmark specifically designed to enable direct comparison between pinhole and fisheye cameras in the same driving environments.
	\item We propose two frameworks, FisheyeBEVDet and FisheyePETR, which leverage spherical modeling tailored for fisheye optics to improve performance.
	\item Extensive experiments on Fisheye3DOD validate our findings and demonstrate significant performance gains over direct transfer baselines.
\end{itemize}

\section{Related Work}

\noindent\textbf{Multi-View 3D Object Detection.}
Most existing multi-view 3DOD methods rely on pinhole images as input, which can be broadly divided into two groups~\cite{mao20233d}:

1) \textit{BEV-based methods}~\cite{huang2021bevdet, huang2022bevdet4d, li2023bevdepth, li2023bevstereo, li2024bevnext, li2023fb} construct 3D detection spaces through Lift-Splat-Shoot (LSS)~\cite{philion2020lift}, which lifts 2D features into 3D via depth-context outer products and projects them onto BEV grids. To improve the accuracy of depth estimation, BEVDepth~\cite{li2023bevdepth} and BEVStereo~\cite{li2023bevstereo} introduce explicit supervision and temporal stereo cues, respectively. 
Additionally, some works focus on improving BEV features, such as bidirectional projection compensation in FB-BEV~\cite{li2023fb} and CRF-modulated depth estimation in BEVNeXt~\cite{li2024bevnext,liu2014discrete}.

2) \textit{Query-based methods}~\cite{wang2022detr3d, liu2022petr, li2022bevformer, yang2023bevformer, liu2023petrv2, wang2023exploring} utilize DETR's sparse 3D queries~\cite{carion2020end} projected onto multi-view images for feature sampling, with Transformer layers~\cite{vaswani2017attention} decoding bounding boxes.
Building upon the query-based paradigm, some works~\cite{liu2022petr, liu2023petrv2, wang2023exploring} simplify projection via 3D positional encoding, while others~\cite{li2022bevformer, yang2023bevformer} replace object queries with dense BEV-centric queries that aggregate features through spatio-temporal attention.

\noindent\textbf{Fisheye Dataset.} The majority of existing fisheye datasets suffer from limited availability and accessibility of annotations.
WoodScape\cite{yogamani2019woodscape} and SynWoodScape\cite{sekkat2022synwoodscape} pioneered multi-task learning for fisheye imagery. However, WoodScape does not provide 3D annotations, and SynWoodScape, although it includes them, releases only about 500 samples, which is insufficient for training modern detectors.
Real-world datasets such as FisheyeCityscapes\cite{ye2020universal} and Fisheye8K~\cite{gochoo2023fisheye8k} are limited by task-specific constraints, with the former focusing on semantic segmentation and the latter designed for object detection.
For synthetic alternatives, OmniScape~\cite{sekkat2020omniscape} caters to stereo semantic segmentation, whereas Synthetic Urban/Deep360~\cite{won2019sweepnet, li2022mode} target omnidirectional depth estimation.
Moreover, recent benchmarks~\cite{samani2023f2bev, yogamani2024daf, yogamani2024fisheyebevseg} are tailored for BEV semantic segmentation.
Overall, these datasets either lack sufficient 3D annotations, are not publicly available due to commercial constraints, or do not simultaneously provide surround-view pinhole and fisheye data necessary for investigating our \textbf{RQ1}.

\begin{table*}[htbp]
	\centering
	{\small
	\begin{tabular}{lccc|ccc|ccc}
		\toprule
		\multirow{2}{*}{\textbf{Dataset}} & \multirow{2}{*}{\textbf{Type}} & \multirow{2}{*}{\textbf{Scenes}} & \multirow{2}{*}{\textbf{Available}} &
		\multicolumn{3}{c|}{\textbf{Image Properties}} & \multicolumn{3}{c}{\textbf{Annotation Details}} \\
		&  &  & & Lens & Frames & Resolution & Class & Anno. & 3D boxes \\
		\midrule
		nuScenes & Real & 1000 & \ding{51} & \textbf{P} & 1.4M & $1600\times900$ & 23 & 40k & 1.4M \\
		Waymo Open & Real & 1150 & \ding{51} & \textbf{P} & 1M & $1920\times1080$ & 4 & 200k & 12M \\
		WoodScape & Real & N/A & no 3D & \textbf{F} & 10k & $1280\times966$ & 3 & 10k & N/A \\
		SynWoodScape & Sim. & N/A & 500 samples & \textbf{F} & 10k & $1280\times966$ & 3 & 10k & N/A \\
		Fisheye8k & Real & 22 & no 3D & \textbf{F} & 8k & $1280\times1280$ & 5 & 8k & \ding{55} \\
		Cognata & Sim. & 5 & \ding{55} & \textbf{F} & 50k & $1920\times1208$ & 5 & 12k & \ding{55} \\
		\midrule
		\multirow{2}{*}{Fisheye3DOD (Ours)} & \multirow{2}{*}{Sim.} & \multirow{2}{*}{8} & \multirow{2}{*}{\ding{51}} & \textbf{P} & 432k & $1280 \times 720$ & \multirow{2}{*}{6}  & \multirow{2}{*}{72k}  & \multirow{2}{*}{607k} \\
		& & & & \textbf{F} & 288k & $800 \times 800$ & & & \\
		\bottomrule
	\end{tabular}}
	\caption{A summary of the proposed Fisheye3DOD dataset and other published datasets, where Bold letters \textbf{P} and \textbf{F} denote pinhole and fisheye images, respectively.}
	\label{tab:dataset}	
\end{table*}

\noindent\textbf{Monocular Fisheye 3D Object Detection.} 
To our knowledge, the work by Plaut et al.~\cite{plaut20213d} is the only study that addresses 3D object detection using fisheye images.  
Their method warps fisheye inputs into cylindrical projections that resemble perspective views, allowing monocular detectors trained on pinhole data to be applied. 
However, it operates on single-view inputs without modeling multi-view feature interactions and does not report performance using pinhole images under the same environment. As a result, it does not address \textbf{RQ1}.

\begin{figure}[tbp]
	\centering
	\includegraphics[width=1\linewidth]{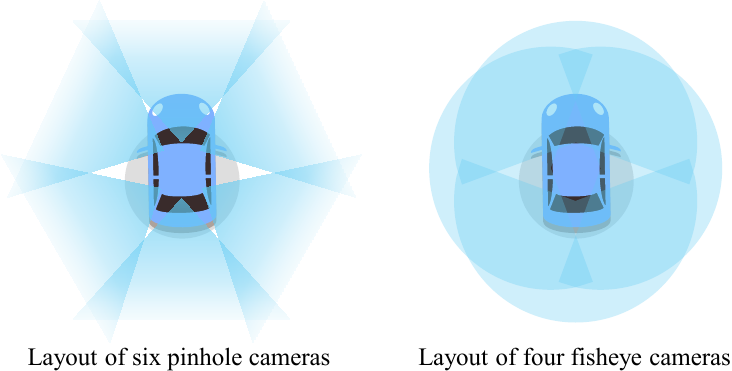}
	\caption{Camera layouts for surround-view perception.}
	\label{fig:sensor}
\end{figure}

\section{Fisheye3DOD Dataset}
\subsection{Data Collection}
To address the existing gap in fisheye 3D object detection datasets, we developed Fisheye3DOD, a synthetic benchmark created through the CARLA simulator~\cite{dosovitskiy2017carla}. 
This dataset comprises 144 driving sequences covering urban and suburban environments, spanning diverse illumination conditions (daytime noon, sunset, night) and weather patterns (clear, cloudy, rainy). 
Each scenario contains temporally aligned sensor data captured at 10Hz over 50-second episodes, yielding a total of 500 frames per sequence.
A detailed comparison between Fisheye3DOD and the existing dataset can be found in Table~\ref{tab:dataset}.

Our sensor configuration includes six surround-view pinhole cameras and four wide-angle fisheye cameras (\( \rm FoV=220^\circ \))~\cite{won2019sweepnet} with corresponding 3D bounding box annotations. 
Moreover, we capture LiDAR point clouds, semantic LiDAR data, and high-precision ego-vehicle trajectories to support potential future work~\cite{huang2023tri, hu2023planning}. 
The sensor pose of Fisheye3DOD adheres to industry-standard practices~\cite{caesar2020nuscenes, won2019sweepnet}, ensuring benchmark compatibility, as shown in Figure~\ref{fig:sensor}. 

Noting that CARLA lacks native fisheye sensor support~\cite{dosovitskiy2017carla}, we mathematically model fisheye distortion via the Kannala-Brandt projection~\cite{kannala2006generic}. 
This projection model captures the non-linear relationship between incident angle \( \theta \) and radial displacement \( r \) through a ninth-order polynomial:
\begin{equation}
	r(\theta) = k_{0}\theta + k_{1}\theta^{3} + k_{2}\theta^{5} + k_{3}\theta^{7} + k_{4}\theta^{9}
	\label{eq:fisheye}
\end{equation}
where \( k=\left\{k_{i}\right\}_{i=0}^{4} \) are commonly used as fisheye-specific distortion coefficients.

\begin{figure}[tbp]
	\centering
	\begin{minipage}{0.58\linewidth}
		\centering
		\includegraphics[width=\linewidth]{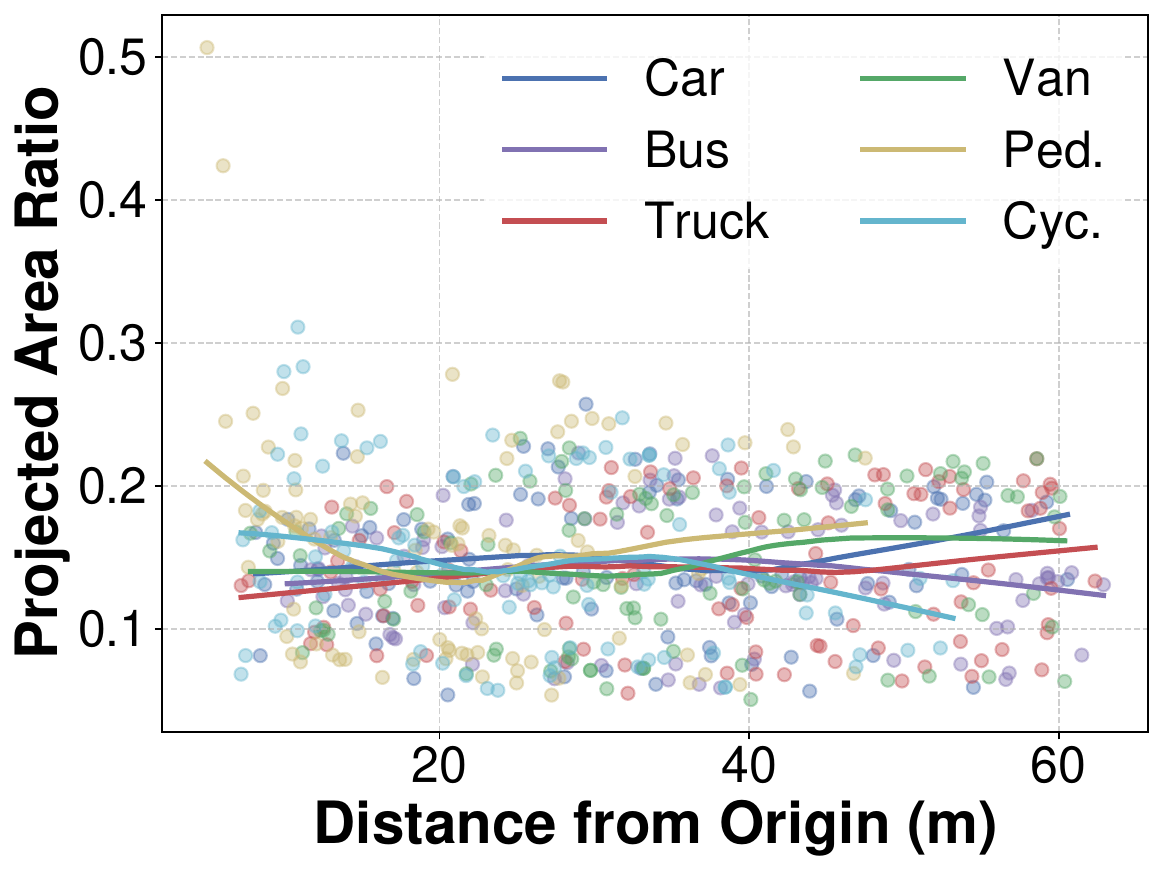}
	\end{minipage}
	\hfill
	\begin{minipage}{0.36\linewidth}
		\centering
		\includegraphics[width=\linewidth]{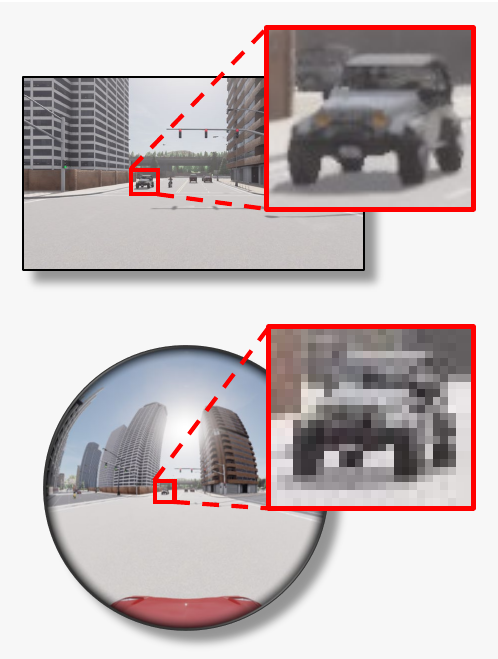}
	\end{minipage}
	\caption{\textbf{Left:} The horizontal axis indicates the 3D distance from the ego vehicle. The vertical axis indicates the ratio between the largest projected 2D bounding box size of an object in any fisheye camera and that in any pinhole camera. The points in the figure correspond to 100 samples per category, with the curves fitted using LOWESS~\cite{cleveland1979robust}. \textbf{Right:} Illustration of pixel compression in fisheye and pinhole images. The same object occupies approximately \( 70 \times 80 \) pixels in the pinhole image, but only about \( 22 \times 26 \) pixels in the fisheye image. The pixel area in the fisheye image is roughly 0.1 times that of the pinhole image.}
	\label{fig:compression}
\end{figure}

\begin{figure*}[tbp]
	\centering
	\includegraphics[width=1\linewidth]{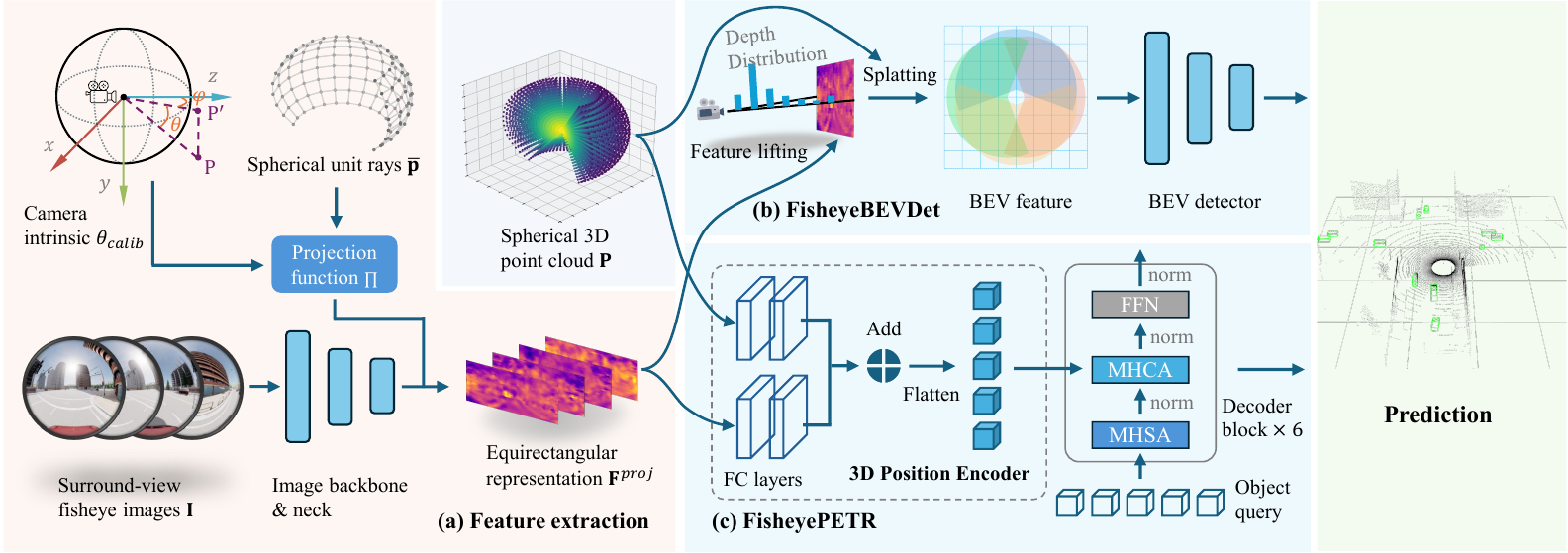}
	\caption{The architecture of the proposed methods. \textbf{(a):} Multi-view fisheye images are processed by a shared backbone, and their features are projected into an equirectangular representation via the projection function \( \Pi \). \textbf{(b):} In FisheyeBEVDet, the projected 2D features are lifted onto a 3D spherical grid to construct a BEV representation. \textbf{(c):} In FisheyePETR, the 2D features are encoded with spherical coordinates and interact with object queries through multi-head cross-attention (MHCA).
	}
	\label{fig:overview}
\end{figure*}

\subsection{Challenges of Fisheye Images}
\label{sec:challenge}
The Figure~\ref{fig:compression} highlights a key challenge in fisheye imagery: pixel compression.
Fisheye projection nonlinearly compresses wide-angle scenes into limited image regions, resulting in significantly fewer pixels per object compared to pinhole projection.
In our dataset, objects in fisheye views occupy only about 15\% of the pixel area of their counterparts in pinhole images, as shown in Figure~\ref{fig:compression} (\textbf{Left}).
This leads to a substantial loss of spatial resolution and visual detail, making reliable detection more challenging.
Note that this loss of information entropy during imaging is irreversible and cannot be recovered through fisheye rectification.
As shown in Figure~\ref{fig:compression} (\textbf{Right}), even when magnified to the same scale, the object region in the fisheye image appears markedly blurrier than that in the pinhole, since no new information is introduced and the original pixel density is limited.

\subsection{Evaluation Metrics}
Our evaluation follows the nuScenes benchmark~\cite{caesar2020nuscenes} protocol, employing center-distance-based AP with decomposed True Positive (\( \mathbb{TP} \)) metrics: mATE (mean Average Translation Error), mASE (mean Average Scale Error), and mAOE (mean Average Orientation Error), to ensure standardized assessment.
This aims to address the shortcomings of IoU-based metrics, where small-footprint objects receive a score of zero even under minor localization errors.
The composite performance metric, named Fisheye Detection Score (FDS), is formulated as:
\begin{equation}
	\mathrm{FDS} = \dfrac{1}{6} \left[3 \mathrm{mAP} + \sum_{\mathrm{mTP} \in \mathbb{TP}} \left( 1 - \mathrm{min} \left( 1, \mathrm{mTP} \right) \right) \right]
	\label{eq:fds}
\end{equation}
Given matching thresholds \( \mathbb{D} = \left\{ 0.5,1,2,4 \right\} \) meters, the mean Average Precision (mAP) across all classes \( \mathbb{C} \) is calculated as \( \mathrm{mAP}=\dfrac{1}{|\mathbb{C}||\mathbb{D}|} \sum\limits_{c\in \mathbb{C}} \sum\limits_{d\in \mathbb{D}} \mathrm{AP_{c,d}}  \).

\section{Our Detector}
\subsection{Spherical Feature Representation}
To fairly answer \textbf{RQ2}, we forgo the ``bells and whistles'' but instead pursue an end-to-end ``tabula rasa'' approach to investigate two distinct 3D detection paradigms: BEV-based and query-based.
Both methods model 3D space in spherical coordinates and perform back-projection to establish image-to-space correspondence.

As illustrated in Figure~\ref{fig:overview}, given a set of surround-view fisheye images \( \mathbf{I}={\{ \mathbf{I}_{i}\in \mathbb{R}^{H_{\mathbf{I}} \times W_{\mathbf{I}} \times 3} \} }_{i=1}^{N} \), each image is directly fed into the backbone network (e.g., ResNet~\cite{he2016deep}) to extract 2D features \( \mathbf{F}^{2d} \).
The extracted features are then warped into an equirectangular representation via the calibrated fisheye projection function \( \Pi \) to align with the spherical 3D space.
Formally, the projected feature map \( \mathbf{F}^{proj} \in \mathbb{R}^{H\times W\times C} \) is computed as:
\begin{equation}
	\mathbf{F}^{proj} = \mathbf{F}^{2d} \circ  \mathbf{G}_{sph}
	\label{eq:proj}
\end{equation}
where \( \circ \) denotes the differentiable warping operation, and \( \mathbf{G}_{sph} \) is a precomputed sampling grid mapping 3D spherical direction vectors to image-plane coordinates via the projection function \( \Pi \). 
Specifically, the grid is derived from the calibrated camera projection model~\cite{scaramuzza2006flexible} as:
\begin{equation}
	\mathbf{G}_{sph} = \sigma\big( \Pi \left( \bar{\mathbf{p}}; \theta_{calib}  \right) \big)
	\label{eq:grid}
\end{equation}
where \( \Pi:\mathbb{R}^3\rightarrow\mathbb{R}^2 \) denotes the projection function from 3D rays to image coordinates, parameterized by the camera calibration parameters \( \theta_{calib} \). The function \( \sigma \) normalizes the coordinates to the range \( [-1, 1] \) for grid sampling. 
Each direction vector \(\bar{\mathbf{p}} \in \mathbb{R}^3\) is parameterized by the azimuth and elevation angles \((\phi, \theta)\) in spherical coordinates as:
\begin{equation}
	\bar{\mathbf{p}} = \left[\cos\theta \cos\phi, \ \sin\theta, \ \cos\theta \sin\phi \right]^\top
\end{equation}
Theoretically, \( \mathbf{F}^{proj} \) can also be represented in cylindrical coordinates with \( \bar{\mathbf{p}} = [\sin\phi, \ y, \ \cos\phi ]^\top \). However, our experiments show that the spherical (equirectangular) representation is superior; please refer to Table~\ref{tab:rq}.

\subsection{BEV-based Detection}
The projected features \( \mathbf{F}^{proj} \) have now been aligned with unit directional vectors on the spherical surface.
We next sample along each \( \bar{\mathbf{p}} \) across discrete depth levels to realize the back-projection from image to 3D space.
Specifically, for FisheyeBEVDet, we represent the BEV space using hierarchical spherical shells, as opposed to the parallel planar stratification in LSS~\cite{philion2020lift}.
Each shell discretizes the depth space aligned with the camera's viewing direction from the ego coordinate origin.

To define these concentric spherical shells, we first discretize the radial depth space into \( D \) uniformly spaced bins ranging from \( r_{min} \) to \( r_{max} \). Let \( r_d \) denote the radial distance at the \( d \)-th depth level:
\begin{equation}
	r_{d} = r_{min} + d \times \delta,\quad d \in \left[0, D - 1\right]_{\mathbb{Z}}
	\label{eq:D}
\end{equation}
where \( \delta=\frac{r_{max}-r_{min}}{D} \) is the fixed interval between consecutive depth levels.
Building upon the defined radial samples, a 3D point at the \( d \)-th shell along the unit direction vector \( \bar{\mathbf{p}} \) corresponding to pixel location \((h,w)\) is computed as:
\begin{equation}
	\mathbf{p}^{cam}_{d,h,w} = r_{d} \times \bar{\mathbf{p}}_{h,w}
	\label{eq:p_cam}
\end{equation}
Each point in the camera coordinate system is then transformed into the unified LiDAR coordinate system using the camera-to-LiDAR transformation matrix \(\mathbf{M} \in \mathbb{R}^{4 \times 4}\):
\begin{equation}
	\mathbf{p}_{d,h,w} = \mathbf{M} \cdot 
	\left[ \left(\mathbf{p}_{d,h,w}^{cam}\right)^\top,\ 1 \right]^\top_{:3}
	\label{eq:p_ref}
\end{equation}
By applying this transformation to all points, we obtain the full point cloud \(\mathbf{P} \in \mathbb{R}^{D \times H \times W \times 3}\).
These points constitute multiple spherical shells that act as spatial anchors for the subsequent BEV feature projection.

Note that for each projected feature \( \mathbf{f}\in\mathbf{F}^{proj} \), its association with a unit ray direction \( \bar{\mathbf{p}} \) on the spherical surface has already been established through the warping process. Based on this, we can estimate a depth probability distribution along the corresponding ray \( \bar{\mathbf{p}} \), enabling compatibility with the LSS paradigm.
Specifically, given the feature \( \mathbf{f} \), a fully-connected (FC) layer predicts a context vector \( \mathbf{c} \in \mathbb{R}^{C} \) and a depth probability distribution \( \alpha \in \Delta^{D-1} \), where \(\Delta^{D-1} := \{\alpha \in \mathbb{R}^D \mid \alpha_d \geq 0, \sum_{d=0}^{D-1} \alpha_d = 1\}\).
The depth-specific feature \( \mathbf{c}_{d} \in \mathbb{R}^{C} \) at \( d \)-th shell is computed as:
\begin{equation}
	\mathbf{c}_{d} = \alpha_{d} \cdot \mathbf{c}
\end{equation}
where \( \alpha_{d} \) denotes the probability of the feature being present at \( d \)-th shell.

Stacking \( \mathbf{c}_{d}  \) across all locations and depths yields a lifted feature volume \( \mathbf{F}^{lift}\in \mathbb{R}^{D \times H \times W \times C} \). 
This volume is then projected into the BEV space using the corresponding 3D points \( \mathbf{P} \), making it compatible with the downstream components of the original BEVDet framework.

\begin{table*}[htbp]
	\centering
	{\small
		\begin{tabular}{llcccccccc}
			\toprule
			\textbf{Methods} & \textbf{Camera} & \textbf{Rectification}  & \textbf{FDS} $\uparrow$ & \textbf{mAP} $\uparrow$ & \textbf{mATE} $\downarrow$ & \textbf{mASE} $\downarrow$ & \textbf{mAOE} $\downarrow$ \\
			\midrule
			BEVDet & 6 $\times$ \textbf{P} & $-$ & 0.563 & 0.506 & 0.458 & 0.161 & 0.520 \\
			BEVDet & 4 $\times$ \textbf{F} & Perspective & 0.440 & 0.304 & 0.588 & 0.177 & 0.505 \\
			BEVDet & 4 $\times$ \textbf{F} & Cylindrical & 0.453 & 0.322 & 0.591 & 0.178 & 0.478 \\
			FisheyeBEVDet & 4 $\times$ \textbf{F} & Cylindrical & 0.476 & 0.361 & 0.581 & 0.162 & 0.482 \\
			FisheyeBEVDet & 4 $\times$ \textbf{F} & Equirectangular & 0.485 & 0.382 & 0.591 & 0.164 & 0.480 \\
			\midrule
			PETR & 6 $\times$ \textbf{P} & $-$ & 0.553 & 0.482 & 0.580 & 0.120 & 0.430 \\
			PETR & 4 $\times$ \textbf{F} & Perspective & 0.408 & 0.274 & 0.783 & 0.161 & 0.433 \\
			PETR & 4 $\times$ \textbf{F} & Cylindrical & 0.411 & 0.285 & 0.773 & 0.169 & 0.447 \\
			FisheyePETR & 4 $\times$ \textbf{F} & Cylindrical & 0.441 & 0.330 & 0.758 & 0.159 & 0.425 \\
			FisheyePETR & 4 $\times$ \textbf{F} & Equirectangular & 0.470 & 0.374 & 0.727 & 0.142 & 0.434 \\
			\bottomrule
	\end{tabular}}
	\caption{Comparison of pinhole-based and fisheye-based methods. Bold letters \textbf{P} and \textbf{F} denote pinhole and fisheye inputs, respectively. Numeric prefixes (e.g., 6 $\times$) indicate camera count. Metric definitions are in the Fisheye3DOD Dataset section.
	}
	\label{tab:rq}
\end{table*}

\begin{table}[htbp]
	\centering
	{\small
	\begin{tabular}{llcc}
		\toprule
		 \textbf{Methods} & \textbf{Camera}   & \textbf{FDS} $\uparrow$ & \textbf{mAP} $\uparrow$ \\
		\midrule
		 BEVDet & 4 $\times$ \textbf{P}~(w/o $\updownarrow$) & 0.370 & 0.206 \\
		 FisheyeBEVDet & 2 $\times$ \textbf{F}~($\updownarrow$)  & 0.454 & 0.324 \\
		 FisheyeBEVDet & 2 $\times$ \textbf{F}~($\leftrightarrow$)  & 0.431 & 0.315 \\
		 FisheyeBEVDet & 4 $\times$ \textbf{F} & 0.485 & 0.382 \\
		\midrule
		PETR & 4 $\times$ \textbf{P}~(w/o $\updownarrow$)& 0.321 & 0.142\\
		FisheyePETR & 2 $\times$ \textbf{F}~($\updownarrow$)  & 0.421 & 0.289\\
		FisheyePETR & 2 $\times$ \textbf{F}~($\leftrightarrow$) & 0.382 & 0.244\\
		FisheyePETR & 4 $\times$ \textbf{F} & 0.470 & 0.374 \\
		\bottomrule
	\end{tabular}}
	\caption{Evaluation of robustness under sensor failure and comparison across different fisheye camera layouts. Arrow directions ($\updownarrow$, $\leftrightarrow$) indicate front-rear and left-right sensor arrangements, respectively.}
	\label{tab:robu}
\end{table}

\subsection{Query-based Detection}
As illustrated in Figure~\ref{fig:overview}(c), FisheyePETR directly encodes the projected features \( \mathbf{F}^{proj} \) with spherical coordinates and leverages object queries to interact with these features via MHCA, enabling end-to-end 3D object detection without explicit BEV representation.

Similar to FisheyeBEVDet, FisheyePETR is also required to construct spherical frustum points as spatial anchors to associate the projected features \( \mathbf{F}^{proj} \) with their corresponding 3D locations.
To align with PETR~\cite{liu2022petr}, FisheyePETR adopts quadratically increasing depth spacing instead of uniform intervals. The depth value \( r_d \) at the \( d \)-th level is computed as:
\begin{equation}
	r_{d}=r_{min}+\frac{r_{max}-r_{min}}{D(D+1)}\times d(d+1)
\end{equation}
Based on these depths, the 3D spherical frustum points are computed by combining each \( r_d \) with the corresponding unit directional vectors using the geometric mappings defined in Eq.\ref{eq:p_cam} and Eq.\ref{eq:p_ref}. These points are then used as positional encodings, which are fused with the projected features \( \mathbf{F}^{proj} \) to enhance 3D spatial awareness in subsequent processing.

Finally, FisheyePETR employs a detection transformer decoder, where object queries interact through self-attention and attend to the projected features via cross-attention, enabling end-to-end 3D detection.

\section{Experiments}
\subsection{Implementation Details}
The experiments are implemented on a single NVIDIA A6000 GPU platform.
The dataset is split into training and testing sets based on scene sequences, using the first 70\% of frames from each scene for training and the remaining 30\% for testing.
Similar to nuScenes~\cite{caesar2020nuscenes}, the Fisheye3DOD dataset is sampled at 2Hz intervals throughout the experiments.
The model is trained for 20 epochs with a batch size of 4. The AdamW optimizer~\cite{loshchilov2017decoupled} is employed for parameter updates, configured with an initial learning rate of 0.0002 and weight decay of 0.01.
The learning rate first undergoes a linear warm-up for the first 500 iterations, followed by a cosine annealing schedule~\cite{loshchilov2016sgdr}.
The detection range spans a cuboid volume, defined by the bounds \( \{ (X, Y, Z) \mid X \in [-48, 48] \, \text{m}, Y \in [-48, 48] \, \text{m}, Z \in [-5, 5] \, \text{m} \} \). 
During the training phase, the dataset is loaded through a class-balanced sampler (CBGS~\cite{zhu2019class}) for effective mitigation of potential data imbalance.

\subsection{Answering the Research Questions}
Here, we evaluate the research questions posed earlier. The experimental results are presented in Table~\ref{tab:rq}.

\textbf{RQ1: How much accuracy is lost when transferring pinhole-based detectors to fisheye images?}
Table~\ref{tab:rq} presents the performance of representative pinhole-based 3D object detectors applied to fisheye data after standard rectification using perspective or cylindrical projection.
Despite this preprocessing, both BEVDet and PETR suffer substantial accuracy drops compared to their original configurations on 6-camera pinhole images. Specifically, FDS decreases by over 12 points for both models, and other metrics also show clear degradation.
This degradation stems from the intrinsic limitations of fisheye imaging. 
As previously discussed in the fisheye challenges, due to nonlinear projection compression, objects in fisheye images occupy only about 15\% of the pixel area they would in pinhole images. 
This severe reduction in effective pixel density leads to irreversible information loss, which cannot be recovered through rectification and directly impacts detection performance.

\textbf{RQ2: How can the transfer be made more effective?}
Table~\ref{tab:rq} compares our proposed methods, FisheyeBEVDet and FisheyePETR, against fisheye-input baselines which rely on image-level rectification, including both perspective and cylindrical projections. 
Both methods consistently outperform their rectified baselines by a significant margin across all key metrics. 
In particular, FisheyeBEVDet and FisheyePETR with equirectangular representation improve FDS by 4.5 and 6.2 points over the perspective-rectified baseline.
This improvement stems from end-to-end modeling of fisheye geometry at the feature level, which preserves richer spatial and semantic information than image-level rectification.
Moreover, they outperform their cylindrical counterparts by 0.9 and 2.9 points, respectively.
This may be due to their more uniform angular sampling, particularly along the vertical direction. 

It should be acknowledged that, despite these improvements, fisheye-based methods still lag behind pinhole detectors due to intrinsic imaging challenges.
In our dataset, pinhole images provide nearly ten times the effective pixel area of fisheye images for objects.
It is therefore unrealistic to expect fisheye-based detectors to achieve comparable accuracy while operating with significantly less spatial evidence.

\subsection{Additional Analysis}
To gain deeper insights, we conduct additional analysis and identify the following key Research Findings (\textbf{RFs}):

\textbf{RF1 (System Robustness): Multi-fisheye systems inherently mitigate sensor failures via extensive FoV overlap.}	
Recent works~\cite{ge2023metabev, yan2023cross, xie2025benchmarking} have explored 3D detection under partial sensor failures. 
We argue that the extensive FoV overlap inherent in multi-fisheye setups naturally provides strong robustness against such failures.
To validate this, Table~\ref{tab:robu} compares pinhole and fisheye configurations both missing front and rear cameras: specifically, BEVDet with four pinhole cameras without front-rear sensors (4 $\times$ \textbf{P} (w/o $\updownarrow$)) versus FisheyeBEVDet with two fisheye cameras arranged left-right (2 $\times$ \textbf{F} ($\leftrightarrow$)).
Similar comparisons are made for PETR variants.
Results show that fisheye methods experience much smaller drops in FDS than pinhole counterparts when front and rear cameras are removed. This is because pinhole setups develop blind spots under these extreme conditions, while multi-fisheye setups maintain full coverage.

\textbf{RF2 (Sensor Layout Impact): Front-rear sensor layouts outperform lateral ones, with full surround achieving the best results.}
To assess the effect of sensor placement in multi-fisheye systems, 
we compare three multi-fisheye layouts: front-rear, left-right, and full surround.
As shown in Table~\ref{tab:robu}, front-rear yields 2-4\% higher FDS than left-right, since most traffic participants (e.g., cars, vans) cluster along the vehicle's longitudinal axis, allowing better shape preservation by reducing radial distortion. In contrast, the left-right layout pushes objects toward image edges, increasing pixel compression and detection difficulty.
Moreover, the full surround layout further improves FDS by 3-5\% over front-rear, achieving the highest accuracy. This improvement arises from multi-camera synergy: front-rear sensors optimize longitudinal coverage, while full surround mitigates lateral distortion by preserving complementary edge details.

\begin{table}[tbp]
	\centering
	{\small
	\begin{tabular}{lcccc}
		\toprule
		\multirow{2}{*}{\textbf{Method}} & \multicolumn{2}{c}{\textbf{0-30 m}}  & \multicolumn{2}{c}{\textbf{0-48 m}} \\
		& \textbf{FDS} $\uparrow$ & \textbf{mAP} $\uparrow$ & \textbf{FDS} $\uparrow$ & \textbf{mAP} $\uparrow$ \\
		\midrule
		BEVDet  & 0.673 & 0.684 & 0.563 & 0.506 \\
		FisheyeBEVDet & 0.586 & 0.555 & 0.485 & 0.382 \\
		\midrule
		PETR  & 0.652 & 0.634 & 0.553 & 0.482 \\
		FisheyePETR  & 0.564 & 0.516 & 0.470 & 0.374 \\
		\bottomrule
	\end{tabular}}
	\caption{Detection performance (FDS and mAP) across cumulative distance ranges (0-30 m to 0-48 m).}
	\label{tab:distance}
\end{table}

\begin{figure*}[htbp]
	\centering
	\includegraphics[width=0.95\linewidth]{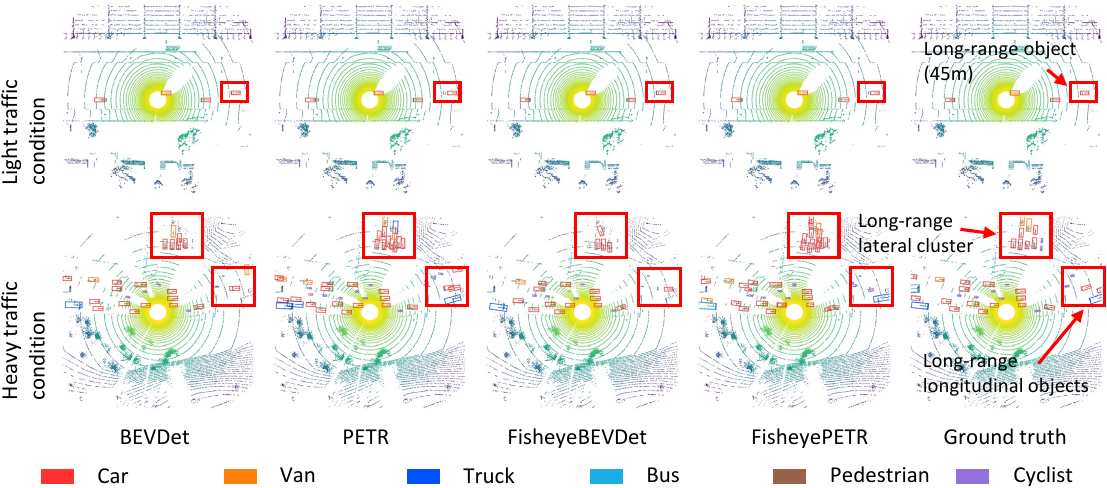}
	\caption{We visualize the predictions in LiDAR point clouds for a clearer comparison. The first row shows a sparse traffic scenario with isolated vehicles on open roads, whereas the second row depicts a dense urban junction with multi-object occlusion.}
	\label{fig:qual}
\end{figure*}

\textbf{RF3 (Distance Degradation): Fisheye camera capabilities align well with near-field sensing.}
Recent studies advocate for the use of fisheye cameras in near-field perception. 
For example, F2BEV~\cite{samani2023f2bev} and FisheyeBEVSeg~\cite{yogamani2024fisheyebevseg, yogamani2024daf} adopt perception ranges of only 16 and 25 meters, respectively.
Our experiments in Table~\ref{tab:distance} validate this, showing fisheye variants achieve 0.586 FDS at 0-30m, comparable to pinhole systems' 0.563 FDS at 0-48m — a critical range covering the under-30m braking distance at 60 km/h~\cite{hosseinlou2012study}. This capability makes fisheye cameras especially suitable for low-speed scenarios such as automated parking systems, warehouse robots, and sidewalk delivery robots.

\begin{figure}[tbp]
	\centering
	\includegraphics[width=1\linewidth]{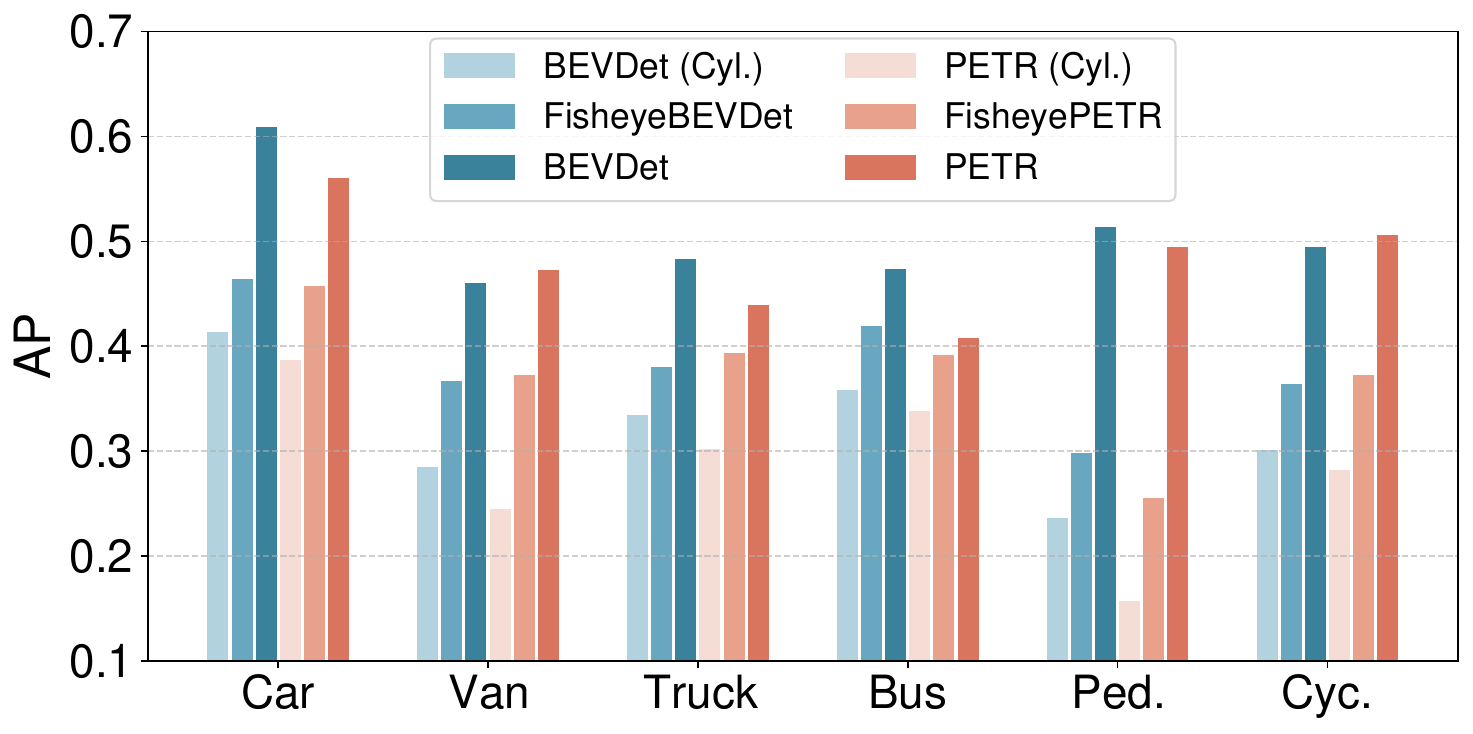}
	\caption{Per-class detection performance. Models with (Cyl.) use cylindrical rectified fisheye images as input.}
	\label{fig:ap_class}
\end{figure}

\textbf{RF4 (Failure Modes and Limitations): Small-footprint objects exacerbate challenges under fisheye distortion.}
Figure~\ref{fig:ap_class} compares per-class AP between our fisheye models, cylindrical rectified baselines, and their pinhole counterparts.
We observe that small-footprint classes, such as \textit{Pedestrian} and \textit{Cyclist}, suffer the most significant performance drop when shifting input from pinhole to fisheye.
This may be due to their inherently small size, which results in fewer visual cues being preserved under fisheye-induced pixel compression.
In simulation environments, these issues are further compounded due to the limited texture richness.
Mitigating this issue may benefit from insights in small object detection.
Notably, our fisheye models significantly outperform their cylindrical baselines across all categories, suggesting the effectiveness of our approach.

\subsection{Qualitative Analysis}
Figure~\ref{fig:qual} presents the prediction visualizations of multiple detectors under varying traffic densities. 
In the light traffic scenario, fisheye variants achieve distance-equivariant detection performance to their standard pinhole counterparts for frontally distant objects (\( \approx \) 45m range), even with their inherent radial distortion and pixel compression.
Under heavy traffic with multi-vehicle occlusion,
all detectors exhibit performance degradation on distant objects, with fisheye variants showing a slightly greater decline. 
This observed discrepancy may be due to the amplified impact of pixel compression under dense occlusion.
Notably, fisheye variants maintain near-field detection accuracy equivalent to pinhole models even under these challenging conditions, suggesting their potential as complementary sensors for close-range perception tasks.

\section{Conclusion}
We present Fisheye3DOD, a benchmark dataset featuring synchronized multi-fisheye images and 3D annotations, to enable systematic study of fisheye-based 3D detection. Based on this dataset, we develop FisheyeBEVDet and FisheyePETR, two end-to-end multi-view detectors tailored for fisheye imagery. Our best model outperforms rectification baselines by up to 6.2 FDS. While a performance gap remains compared to pinhole systems, fisheye-based detection proves highly suitable for compact, low-speed robotic platforms with strict space constraints.

\section{Acknowledgments}
This research was supported by the Guangzhou Basic and Applied Basic Research Foundation under Grant SL2024A04J0183, the Guangxi Key Research and Development Project under Grant 2024AB08049, the National Natural Science Foundation of China under Grant 92470202, the Fund of National Key Laboratory of Multispectral Information Intelligent Processing Technology (No. 202410487201), and the Major Key Project of Pengcheng Laboratory (PCL2025A02).

\bibliography{aaai2026}

\end{document}